\documentclass{article}

\usepackage{amsmath}
\usepackage{graphicx}
\usepackage[preprint]{neurips_2025}

\usepackage[utf8]{inputenc} 
\usepackage[T1]{fontenc}    
\usepackage{hyperref}       
\usepackage{url}            
\usepackage{booktabs}       
\usepackage{amsfonts}       
\usepackage{nicefrac}       
\usepackage{microtype}      
\usepackage{xcolor}         
\usepackage{booktabs,longtable,array}
\newcolumntype{P}[1]{>{\raggedright\arraybackslash}p{#1}}

\title{RADAR: Mechanistic Pathways for Detecting Data Contamination in LLM Evaluation}

\author{%
Ashish Kattamuri$^{1}$ \quad Harshwardhan Fartale$^{2}$ \quad Arpita Vats$^3$ \quad Rahul Raja$^3$ \quad Ishita Prasad$^4$ \\
$^1$Proofpoint \quad $^2$Indian Institute of Science \quad $^3$ Linkedin \quad $^4$ Meta FAIR\\
}

\begin{document}

\maketitle

\begin{abstract}
Data contamination poses a significant challenge to reliable LLM evaluation, where models may achieve high performance by memorizing training data rather than demonstrating genuine reasoning capabilities. We introduce RADAR (Recall vs. Reasoning Detection through Activation Representation), a novel framework that leverages mechanistic interpretability to detect contamination by distinguishing recall-based from reasoning-based model responses. RADAR extracts 37 features spanning surface-level confidence trajectories and deep mechanistic properties including attention specialization, circuit dynamics, and activation flow patterns. Using an ensemble of classifiers trained on these features, RADAR achieves 93\% accuracy on a diverse evaluation set, with perfect performance on clear cases and 76.7\% accuracy on challenging ambiguous examples. This work demonstrates the potential of mechanistic interpretability for advancing LLM evaluation beyond traditional surface-level metrics. The code used in this work is publicly available\footnote{\url{https://colab.research.google.com/drive/1Bio-yt2rdoo4ODX_xGUJqJm1iXNbF2Xy?usp=sharing}}.

\renewcommand{\thefootnote}{\fnsymbol{footnote}}
\footnotetext[1]{This work does not relate to positions at Meta, LinkedIn, Proofpoint or Indian Institute of Science}
\renewcommand{\thefootnote}{\arabic{footnote}}

\end{abstract}

\section{Introduction}

Large Language Models (LLMs) show strong performance across tasks, but data contamination remains a major challenge in evaluation. Overlap between training and evaluation sets inflates metrics and obscures the distinction between genuine reasoning and memorization \citep{golchin2023time, deng2023investigating, feldman2020does}.

Existing detection methods typically compare evaluation data to training corpora, check n-gram overlaps, or flag verbatim outputs \citep{carlini2021extracting}. These approaches are limited: they require access to training data, fail with paraphrased contamination, and cannot reveal whether a model solved a task by recall or reasoning.

We propose \textbf{RADAR}, which instead analyzes internal computation dynamics. Leveraging mechanistic interpretability, RADAR extracts features from attention, hidden states, and activation flows \citep{elhage2021mathematical, olah2020zoom}. Recall exhibits focused attention and rapid confidence convergence, while reasoning shows distributed activation and gradual stabilization.

Our contributions are: (1) We demonstrate that mechanistic features can reliably distinguish recall from reasoning with 93\% accuracy, (2) We provide interpretable insights into the internal signatures of these cognitive processes, and (3) We offer a practical tool for contamination detection that works without access to training data.

\section{Methodology}
\subsection{Framework Architecture}
\begin{figure}[h!]
\centering
\includegraphics[width=1.0\textwidth]{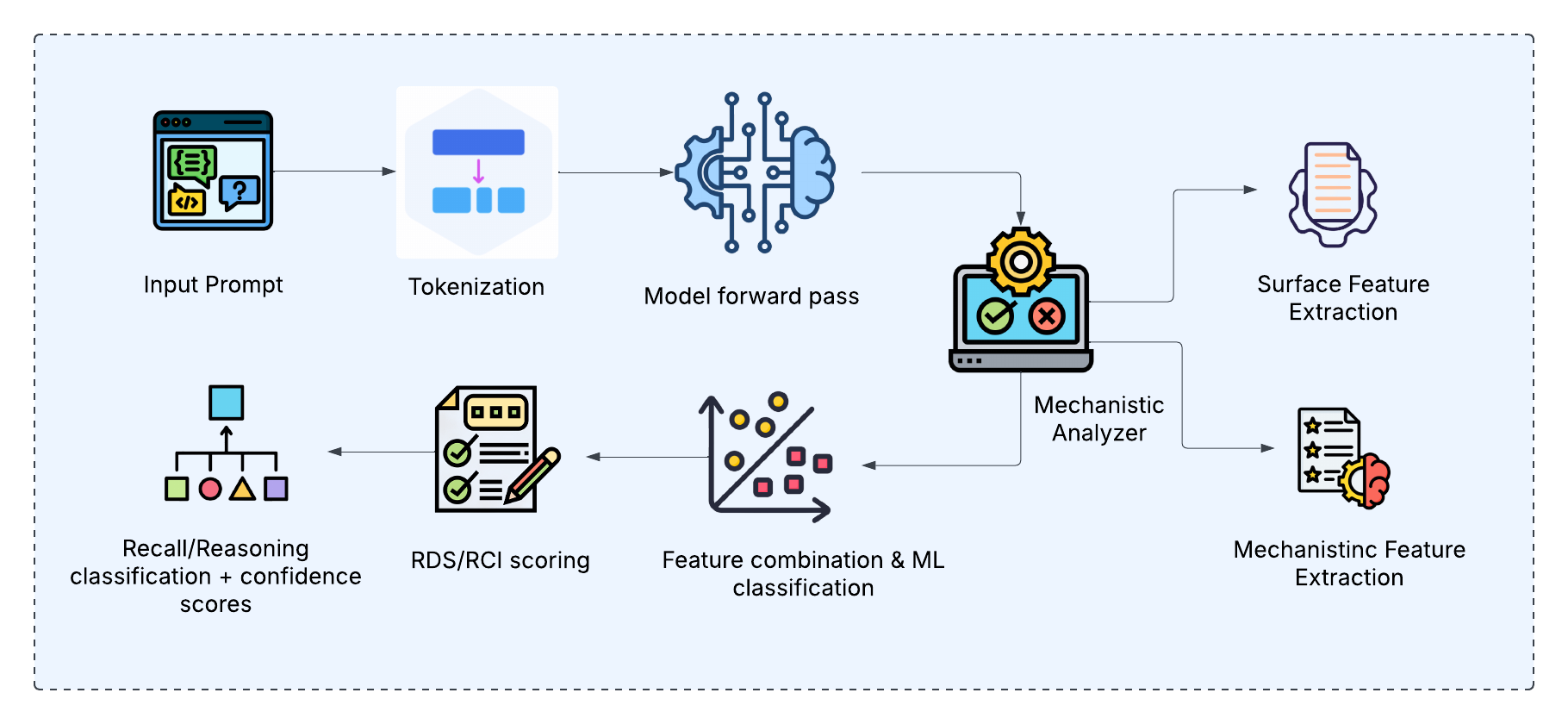}
\caption{RADAR Framework Architecture: Input prompts are processed by the Mechanistic Analyzer to extract internal states, which are converted to Surface and Mechanistic Features, then classified by an ensemble to predict recall vs. reasoning with confidence scores.}
\label{fig:architecture}
\end{figure}

RADAR operates through three integrated components: (1) \textbf{Mechanistic Analyzer} that extracts internal model states, (2) \textbf{Feature Extraction} that computes surface and mechanistic features, and (3) \textbf{Classifier} that predicts recall vs. reasoning, , as illustrated in Figure~\ref{fig:architecture}..

The Mechanistic Analyzer interfaces with target LLMs (e.g., DialoGPT-medium) configured to output attention weights and hidden states. For each prompt, it analyzes attention patterns across all heads and layers, computing entropy and specialization metrics, and examines hidden state dynamics, including variance, norms, and effective rank.

\subsection{Feature Engineering}

We extract 37 features organized into two complementary categories:

\textbf{Surface Features (17):} Derived from the model's output trajectory across layers, these features capture prediction dynamics through confidence statistics (mean, std, max, min, range), convergence properties (layer, speed, slope), entropy measures (mean, change, information gain), and stability metrics.

\textbf{Mechanistic Features (20):} Derived from attention weights and hidden states across all layers and heads, these features capture internal computational mechanisms, including attention specialization (specialized heads, specialization scores, entropy), circuit dynamics (depth, complexity, activation flow), intervention sensitivity (ablation robustness, critical components), working memory (hidden state variance, norm trajectories), and causal effects (logit attribution, mediation scores).





\subsection{Classification System}

The classification module employs an ensemble of four supervised learning models: Random Forest\cite{breiman2001random}, Gradient Boosting\cite{friedman2001greedy}, Support Vector Machine (SVM)\cite{cortes1995support}, and Logistic Regression\cite{hosmer2013applied}. Each model is trained on the extracted feature vectors after normalization with \texttt{StandardScaler}. The scaling ensures zero mean and unit variance across features:
\[
x_i' = \frac{x_i - \mu_i}{\sigma_i},
\]
where $\mu_i$ and $\sigma_i$ denote the mean and standard deviation of feature $i$.

For prediction, each base classifier $j \in \{1,\dots,M\}$ (with $M=4$) outputs a hard label $\hat{y}_j$ and a probability estimate $p_j = P(y=1 \mid x')$, where $y=1$ corresponds to \emph{recall} and $y=0$ to \emph{reasoning}. The ensemble aggregates these outputs as:
\[
\hat{y} = 1 \left[ \frac{1}{M} \sum_{j=1}^{M} \hat{y}_j > \frac{1}{2} \right], 
\qquad 
\bar{p} = \frac{1}{M} \sum_{j=1}^{M} p_j
\]

The final confidence score is defined consistently with the predicted label:
\[
\mathrm{conf} = \bar{p}, \; \hat{y}=1 \; (\text{recall}), 
\qquad 
1-\bar{p}, \; \hat{y}=0 \; (\text{reasoning})
\]

\section{Experiments and Results}

\subsection{Experimental Setup and Results}

We curated two datasets: a balanced training set (30 examples: 15 recall, 15 reasoning) and a diverse test set (100 examples: 20 clear recall, 20 clear reasoning, 30 challenging cases, 30 complex reasoning). The classifier achieved 96.7\% cross-validation accuracy during training.

RADAR achieved an overall accuracy of 93.0\% on the test set, with task-specific performance of 97.7\% on recall tasks and 89.3\% on reasoning tasks. A detailed breakdown of performance across different categories is shown in Table~\ref{tab:radar_results}.

\begin{table}[h!]
\centering
\caption{RADAR Performance Results}
\label{tab:radar_results}
\begin{tabular}{c c}
\toprule
\begin{tabular}{l c}
\multicolumn{2}{c}{\textit{Overall Performance}} \\
\midrule
\hspace{2mm} Overall Accuracy & 93.0\% \\
\hspace{2mm} Recall Tasks & 97.7\% \\
\hspace{2mm} Reasoning Tasks & 89.3\% \\
\end{tabular}
&
\begin{tabular}{l c}
\multicolumn{2}{c}{\textit{Category-wise Performance}} \\
\midrule
\hspace{2mm} Clear Recall & 100\% (20/20) \\
\hspace{2mm} Clear Reasoning & 100\% (20/20) \\
\hspace{2mm} Challenging Cases & 76.7\% (23/30) \\
\hspace{2mm} Complex Reasoning & 100\% (30/30) \\
\end{tabular}
\\
\bottomrule
\end{tabular}
\end{table}

\subsection{Feature Analysis}

\begin{figure}[h!]
\centering
\includegraphics[width=\textwidth]{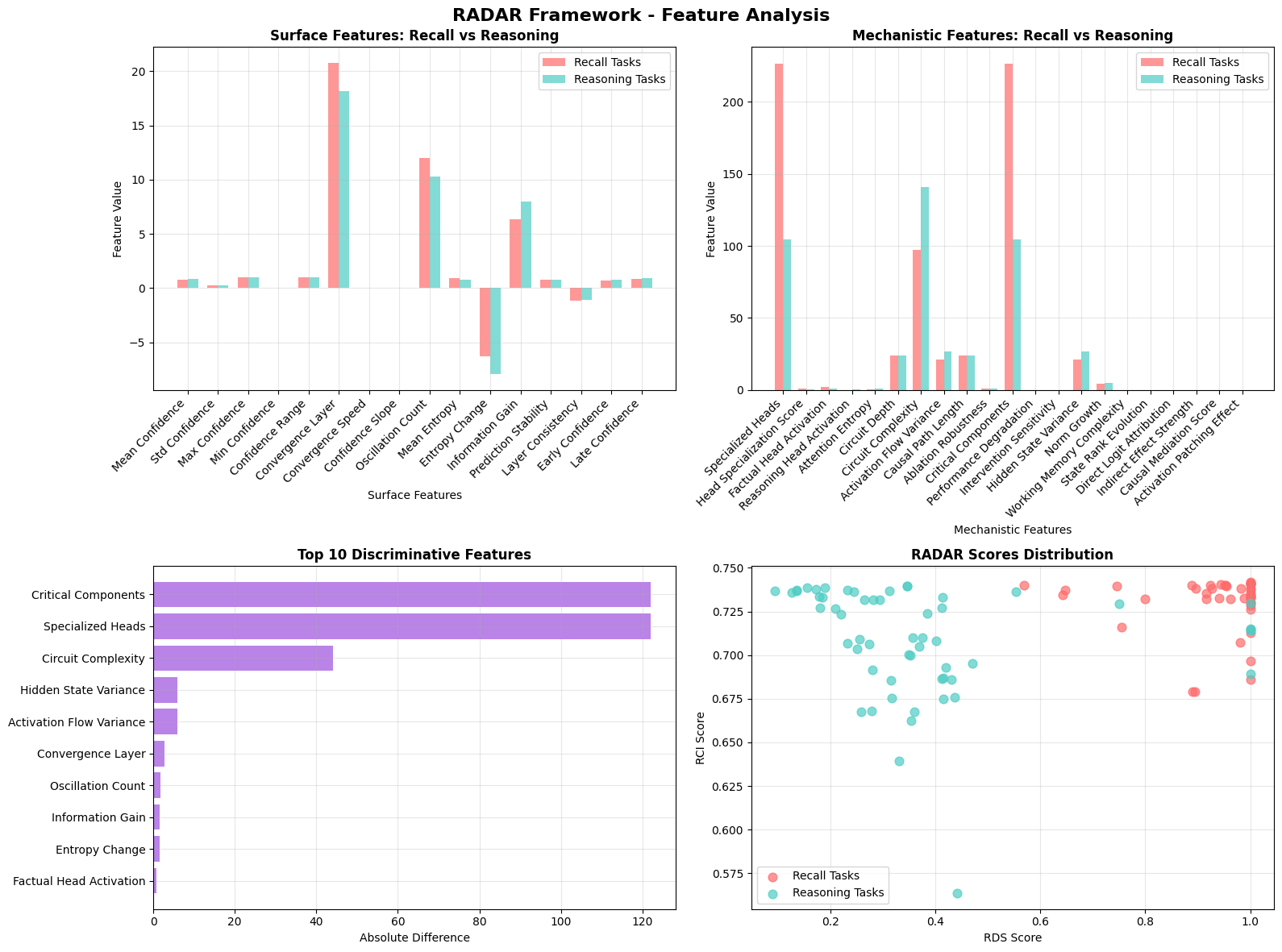}
\caption{RADAR Feature Analysis: Comparison of surface and mechanistic features for recall and reasoning tasks, highlighting top discriminative features and RDS–RCI score distribution. The results show recall tasks characterized by early confidence and specialized heads, while reasoning tasks rely on broader circuit complexity and higher activation flow variance. The scatter plot demonstrates strong clustering, with recall tasks in the high-RDS region and reasoning tasks distributed in lower-RDS regions.}
\label{fig:features}
\end{figure}

Key discriminative features include specialized attention heads (higher for recall), circuit complexity (higher for reasoning), and confidence convergence patterns (faster for recall). Recall tasks showed a mean Recall Detection Score (RDS) of 0.933 compared to 0.375 for reasoning, demonstrating clear separability, as shown in Figure~\ref{fig:features}.

Surface features revealed that recall tasks exhibit higher early confidence and faster convergence, whereas reasoning tasks show gradual confidence build-up and later stabilization. Mechanistic features highlighted that recall relies on focused attention patterns and specialized heads, while reasoning engages broader network resources with higher activation flow variance. The scatter plot visualization confirms clear separation between recall and reasoning tasks in the RDS–RCI score space.

\section{Discussion and Implications}

\subsection{Contamination Detection Applications}

RADAR's ability to distinguish recall from reasoning has direct implications for contamination detection. When reasoning-type prompts elicit recall-like internal signatures (high confidence, fast convergence, specialized attention), this indicates potential contamination where the model "knows" rather than "computes" the answer.

Our approach offers several benefits: (1) Works without access to training data, (2) Analyzes computational processes rather than just outputs, (3) Provides interpretable features explaining classifications, (4) Complements existing external detection methods, and (5) Scales to different model architectures.

\subsection{Interpretability Insights}

The feature analysis confirms that recall and reasoning leave distinct mechanistic signatures. Recall processes exhibit focused attention patterns with rapid confidence convergence, suggesting direct retrieval pathways. Reasoning processes show distributed attention, gradual confidence build-up, and higher circuit complexity, indicating multi-step computational processes engaging broader network resources. Top discriminative features (Specialized Heads, Circuit Complexity, Hidden State Variance) capture fundamental differences in how the model processes information, providing interpretable insights into the underlying cognitive mechanisms.

\section{Conclusion}
RADAR demonstrates that mechanistic interpretability can effectively detect data contamination by analyzing internal LLM processing signatures. Our framework achieves 93\% accuracy in distinguishing recall from reasoning, providing interpretable insights into the cognitive processes underlying model responses. This work opens new directions for LLM evaluation that move beyond surface-level metrics to examine computational mechanisms. The ability to detect contamination without training data access, combined with interpretable mechanistic features, makes RADAR a valuable tool for improving LLM evaluation reliability.

Future work will explore scaling to larger models, developing unsupervised detection methods, and extending to other contamination types. The integration of mechanistic interpretability with traditional evaluation methods promises more robust and trustworthy LLM assessment frameworks.

\bibliographystyle{unsrtnat}
\bibliography{references}

\begin{thebibliography}{10}
\providecommand{\natexlab}[1]{#1}
\providecommand{\url}[1]{\texttt{#1}}
\expandafter\ifx\csname urlstyle\endcsname\relax
  \providecommand{\doi}[1]{doi: #1}\else
  \providecommand{\doi}{doi: \begingroup \urlstyle{rm}\Url}\fi

\bibitem[Golchin and Surdeanu(2023)]{golchin2023time}
S.~Golchin and M.~Surdeanu.
\newblock Time travel in llms: Tracing data contamination in large language models.
\newblock \emph{arXiv preprint arXiv:2308.08493}, 2023.

\bibitem[Deng et~al.(2023)Deng, Zhao, Tang, and Gerstein]{deng2023investigating}
C.~Deng, Y.~Zhao, X.~Tang, and M.~Gerstein.
\newblock Investigating data contamination in modern benchmarks for large language models.
\newblock \emph{arXiv preprint arXiv:2311.09783}, 2023.

\bibitem[Feldman(2020)]{feldman2020does}
V.~Feldman.
\newblock Does learning require memorization? a short tale about a long tail.
\newblock In \emph{Proceedings of the 52nd Annual ACM SIGACT Symposium on Theory of Computing}, pages 954--959, 2020.

\bibitem[Carlini et~al.(2021)Carlini, Tramer, Wallace, Jagielski, Herbert-Voss, Lee, et~al.]{carlini2021extracting}
N.~Carlini, F.~Tramer, E.~Wallace, M.~Jagielski, A.~Herbert-Voss, K.~Lee, et~al.
\newblock Extracting training data from large language models.
\newblock In \emph{30th USENIX Security Symposium (USENIX Security 21)}, pages 2633--2650, 2021.

\bibitem[Elhage et~al.(2021)Elhage, Nanda, Olsson, Henighan, Joseph, Mann, et~al.]{elhage2021mathematical}
N.~Elhage, N.~Nanda, C.~Olsson, T.~Henighan, N.~Joseph, B.~Mann, et~al.
\newblock A mathematical framework for transformer circuits.
\newblock \emph{Anthropic}, 2021.

\bibitem[Olah et~al.(2020)Olah, Cammarata, Schubert, Goh, Petrov, and Carter]{olah2020zoom}
C.~Olah, N.~Cammarata, L.~Schubert, G.~Goh, M.~Petrov, and S.~Carter.
\newblock Zoom in: An introduction to circuits.
\newblock \emph{Distill}, 5\penalty0 (3):\penalty0 e00024--001, 2020.

\bibitem[Breiman(2001)]{breiman2001random}
Leo Breiman.
\newblock Random forests.
\newblock \emph{Machine learning}, 45\penalty0 (1):\penalty0 5--32, 2001.

\bibitem[Friedman(2001)]{friedman2001greedy}
Jerome~H Friedman.
\newblock Greedy function approximation: A gradient boosting machine.
\newblock \emph{Annals of statistics}, pages 1189--1232, 2001.

\bibitem[Cortes and Vapnik(1995)]{cortes1995support}
Corinna Cortes and Vladimir Vapnik.
\newblock Support-vector networks.
\newblock \emph{Machine learning}, 20:\penalty0 273--297, 1995.

\bibitem[Hosmer et~al.(2013)Hosmer, Lemeshow, and Sturdivant]{hosmer2013applied}
David~W. Hosmer, Stanley Lemeshow, and Rodney~X. Sturdivant.
\newblock \emph{Applied Logistic Regression}.
\newblock Wiley, 2013.

\end{thebibliography}

\appendix

\section{Implementation Details}

\subsection{Model Configuration}
The target model (microsoft/DialoGPT-medium) was configured with \texttt{output\_attentions=True} and \texttt{output\_hidden\_states=True}. Analysis focused on input prompt tokens to capture reasoning during comprehension.

\subsection{Feature Computation}
Surface features tracked confidence and entropy trajectories across layers. Mechanistic features analyzed attention weight distributions using entropy measures and computed hidden state statistics including effective rank via SVD decomposition.

\subsection{Training Procedure}
Features were preprocessed and standardized using StandardScaler. Each ensemble model was trained with k-fold cross-validation for robust performance estimation.

\section{Training and Test Datasets}

This appendix provides details on the datasets used for training and evaluating the Enhanced RADAR Framework's Recall--Reasoning Classifier.

\subsection{Training Dataset}

The training dataset was used exclusively for training the classifier. It consists of 30 examples, each containing a prompt and a corresponding label indicating whether the expected response is based on ``recall'' or ``reasoning.'' The composition is as follows:

\begin{table}[h!]
\centering
\caption{Composition of the training dataset.}
\begin{tabular}{l c}
\toprule
\textbf{Category} & \textbf{Count} \\
\midrule
Total Examples & 30 \\
Recall Examples & 15 \\
Reasoning Examples & 15 \\
\bottomrule
\end{tabular}
\end{table}

This dataset provides the classifier with a basic representation of the internal features and patterns that distinguish factual retrieval from logical inference. Representative examples are shown below:

\begin{table}[h!]
\centering
\caption{Sample training dataset prompts and labels.}
\begin{tabular}{p{0.6\textwidth} p{0.2\textwidth}}
\toprule
\textbf{Prompt} & \textbf{Label} \\
\midrule
``The capital of France is'' & recall \\
``If X is the capital of France, then X is'' & reasoning \\
``2 + 2 equals'' & recall \\
``If a triangle has angles 60, 60, and X degrees, then X equals'' & reasoning \\
\bottomrule
\end{tabular}
\end{table}

\subsection{Test Dataset}

The test dataset was used only for evaluating the trained classifier on unseen data. It comprises 100 examples, with broader coverage to assess generalization across different levels of difficulty and ambiguity:

\begin{table}[h!]
\centering
\caption{Composition of the test dataset.}
\begin{tabular}{l c}
\toprule
\textbf{Category} & \textbf{Count} \\
\midrule
Total Examples & 100 \\
Clear Recall Examples & 20 \\
Clear Reasoning Examples & 20 \\
Challenging/Ambiguous Cases & 30 \\
Complex Reasoning Cases & 30 \\
\bottomrule
\end{tabular}
\end{table}

The inclusion of challenging and complex reasoning cases is important for evaluating robustness, especially in detecting possible data contamination where a reasoning task could be solved by recall.

Examples from each category are shown below:

\begin{table}[h!]
\centering
\caption{Sample test dataset prompts by category.}
\begin{tabular}{p{0.2\textwidth} p{0.6\textwidth} p{0.15\textwidth}}
\toprule
\textbf{Category} & \textbf{Example Prompt} & \textbf{Label} \\
\midrule
Clear Recall & ``The capital of Germany is'' & recall \\
Clear Reasoning & ``If a rectangle has length 5 and width 3, its area is'' & reasoning \\
Challenging/Ambiguous & ``What is the sum of 10 and 15?'' & reasoning \\
Complex Reasoning & ``If a store has 100 items and sells 30\% of them, how many items remain?'' & reasoning \\
\bottomrule
\end{tabular}
\end{table}

\subsection{Why Challenging or Ambiguous Prompts Are Difficult}

Challenging or ambiguous prompts are difficult because they blur the line between recall and reasoning.  
\begin{itemize}
    \item Some prompts may appear to require reasoning (e.g., arithmetic) but can be solved by memorized recall if the model has seen similar examples during training.  
    \item Conversely, some factual prompts may trigger reasoning-like processing if the information is incomplete or framed indirectly.  
    \item Ambiguity arises when the surface form of the task does not clearly signal whether the solution requires stored knowledge or active inference.  
\end{itemize}

These cases are crucial for evaluation because they reveal whether the classifier is robust to subtle shifts in task framing and whether it can correctly separate recall-driven answers from reasoning-based ones.

\section{Scoring}
In addition to the binary classification, the RADAR Framework computes several continuous scores that provide a more nuanced perspective:

\begin{itemize}
    \item \textbf{Recall Detection Score (RDS):} Indicates how strongly the analysis suggests a recall-based process, combining specific surface and mechanistic features.
    \item \textbf{Reasoning Complexity Index (RCI):} Reflects the complexity and depth of processing, suggesting a reasoning-based process. Derived from a combination of surface and mechanistic features.
    \item \textbf{Mechanistic Score:} Focuses on features related to causal effects and intervention sensitivity.
    \item \textbf{Circuit Complexity Score:} Based on features describing the depth and complexity of the activated computational graph.
\end{itemize}

These scores are calculated using predefined formulas that weigh different features according to their relevance to recall and reasoning processes. They provide complementary information to the classifier's binary output.

\section{Feature Documentation}
The RADAR (Recall And Deliberative Analysis of Reasoning) Framework extracts 37 features from language model behavior to distinguish between recall-based and reasoning-based tasks. These features are organized into two categories: \textbf{Surface Features} (16 features) that capture observable trajectory patterns, and \textbf{Mechanistic Features} (21 features) that analyze internal model dynamics through attention patterns and activation analysis.

\subsection{Surface Features (16 Features)}

Surface features analyze the confidence and entropy trajectories across all model layers to capture behavioral patterns without requiring deep mechanistic analysis.

\subsubsection{Confidence-Based Features (8 Features)}

\begin{longtable}{p{0.25\textwidth} p{0.15\textwidth} p{0.55\textwidth}}
\toprule
\textbf{Feature} & \textbf{Type} & \textbf{Definition \& Computation} \\
\midrule

\texttt{mean\_confidence} & float & 
Mean confidence across all layers: $\bar{c} = \frac{1}{L} \sum_{l=1}^{L} c_l$ where $c_l$ is the maximum softmax probability at layer $l$, and $L$ is the total number of layers. \\

\texttt{std\_confidence} & float & 
Standard deviation of confidence trajectory: $\sigma_c = \sqrt{\frac{1}{L-1} \sum_{l=1}^{L} (c_l - \bar{c})^2}$. Higher values indicate more variable confidence across layers. \\

\texttt{max\_confidence} & float & 
Maximum confidence achieved: $c_{max} = \max_{l \in [1,L]} c_l$. Indicates peak certainty reached by the model. \\

\texttt{min\_confidence} & float & 
Minimum confidence observed: $c_{min} = \min_{l \in [1,L]} c_l$. Represents lowest certainty point in processing. \\

\texttt{confidence\_range} & float & 
Range of confidence values: $\Delta c = c_{max} - c_{min}$. Measures the span of confidence variation across layers. \\

\texttt{convergence\_layer} & int & 
Layer index where maximum confidence is achieved: $l^* = \arg\max_{l} c_l$. Earlier convergence may indicate simpler recall tasks. \\

\texttt{convergence\_speed} & float & 
Inverse of convergence layer: $v_{conv} = \frac{1}{l^* + 1}$. Higher values indicate faster convergence to high confidence. \\

\texttt{confidence\_slope} & float & 
Linear regression slope of confidence trajectory: $\beta = \frac{\sum_{l=1}^{L} (l - \bar{l})(c_l - \bar{c})}{\sum_{l=1}^{L} (l - \bar{l})^2}$ where $\bar{l} = \frac{L+1}{2}$. Positive slopes indicate increasing confidence. \\

\bottomrule
\end{longtable}

\subsubsection{Trajectory Dynamics Features (4 Features)}

\begin{longtable}{P{0.25\textwidth} P{0.15\textwidth} P{0.55\textwidth}}
\toprule
\textbf{Feature} & \textbf{Type} & \textbf{Definition \& Computation} \\
\midrule
\endfirsthead

\toprule
\textbf{Feature} & \textbf{Type} & \textbf{Definition \& Computation} \\
\midrule
\endhead

\midrule
\multicolumn{3}{r}{\small Continued on next page} \\
\bottomrule
\endfoot

\bottomrule
\endlastfoot

\texttt{oscillation\_count} & int &
Number of sign changes in the discrete confidence derivative.
Let \(\Delta c_l = c_{l+1} - c_l\) for \(l=1,\dots,L-1\).
Then
\[
\text{oscillation\_count} = \#\{\,l \in \{1,\dots,L-2\} : (\Delta c_l)(\Delta c_{l+1}) < 0 \,\},
\]
i.e., consecutive derivatives with opposite sign. Zeros in \(\Delta c_l\) are ignored for sign changes. \\

\texttt{early\_confidence} & float &
Mean confidence in the first half of layers:
\[
c_{\text{early}} = \frac{1}{\lfloor L/2 \rfloor}\sum_{l=1}^{\lfloor L/2 \rfloor} c_l.
\]
Captures initial model certainty. \\

\texttt{late\_confidence} & float &
Mean confidence in the second half of layers:
\[
c_{\text{late}} = \frac{1}{\lceil L/2 \rceil}\sum_{l=\lfloor L/2 \rfloor + 1}^{L} c_l.
\]
Captures final model certainty. \\

\texttt{prediction\_stability} & float &
Inverse of confidence standard deviation:
\[
s_{\text{pred}} = 1 - \sigma_c, \quad
\sigma_c = \sqrt{\frac{1}{L-1}\sum_{l=1}^{L}\bigl(c_l - \bar{c}\bigr)^2},\quad
\bar{c} = \frac{1}{L}\sum_{l=1}^{L} c_l.
\]
Higher values indicate more stable predictions across layers. \\

\end{longtable}

\subsubsection{Information-Theoretic Features (4 Features)}

\begin{longtable}{P{0.25\textwidth} P{0.15\textwidth} P{0.55\textwidth}}
\toprule
\textbf{Feature} & \textbf{Type} & \textbf{Definition \& Computation} \\
\midrule
\endfirsthead

\toprule
\textbf{Feature} & \textbf{Type} & \textbf{Definition \& Computation} \\
\midrule
\endhead

\midrule
\multicolumn{3}{r}{\small Continued on next page} \\
\bottomrule
\endfoot

\bottomrule
\endlastfoot

\texttt{mean\_entropy} & float &
Average entropy across layers:
\[
\bar{H} = \frac{1}{L}\sum_{l=1}^{L} H_l,\quad
H_l = -\sum_{i} p_{l,i}\,\log p_{l,i},
\]
where \(p_{l,i}\) is the probability of token \(i\) at layer \(l\) and \(\log\) is the natural logarithm. \\

\texttt{entropy\_change} & float &
Change from first to last layer:
\[
\Delta H = H_{L} - H_{1}.
\]
Negative values indicate uncertainty reduction. \\

\texttt{information\_gain} & float &
Negative entropy change:
\[
IG = -\Delta H = H_{1} - H_{L}.
\]
Positive values indicate successful uncertainty reduction. \\

\texttt{layer\_consistency} & float &
Inverse of entropy standard deviation:
\[
\text{consistency} = 1 - \sqrt{\frac{1}{L-1}\sum_{l=1}^{L}\bigl(H_l - \bar{H}\bigr)^2}.
\]
Higher values indicate more consistent information processing across layers. \\

\end{longtable}

\subsection{Mechanistic Features (21 Features)}

Mechanistic features analyze internal model dynamics through attention patterns, activation flows, and causal intervention proxies to understand the computational mechanisms underlying different task types.

\subsubsection{Attention Specialization Features (5 Features)}

\begin{longtable}{p{0.3\textwidth} p{0.15\textwidth} p{0.5\textwidth}}
\toprule
\textbf{Feature} & \textbf{Type} & \textbf{Definition \& Computation} \\
\midrule

\texttt{num\_specialized\_heads} & int & 
Total count of attention heads with entropy below a specialization threshold (typically \(\tau=1.5\)):
\[
N_{\text{spec}} = \sum_{l=1}^{L} \sum_{h=1}^{H} \mathbf{1}[H_{l,h} < \tau],
\]
where \(H_{l,h}\) is the entropy of head \(h\) in layer \(l\). \\

\texttt{head\_specialization\_score} & float & 
Normalized specialization measure:
\[
S_{\text{head}} = 1 - \frac{\bar{H}_{\text{attn}}}{3.0},
\]
where \(\bar{H}_{\text{attn}}\) is the mean attention entropy across all heads. Higher values indicate more specialized attention patterns. \\

\texttt{factual\_head\_activation} & float & 
Inverse relationship with attention entropy:
\[
A_{\text{fact}} = \frac{1}{\bar{H}_{\text{attn}} + \epsilon}, \quad \epsilon=10^{-8}.
\]
Higher values suggest factual recall patterns (low entropy, focused attention). \\

\texttt{reasoning\_head\_activation} & float & 
Proportional to attention entropy:
\[
A_{\text{reason}} = \frac{\bar{H}_{\text{attn}}}{3.0}.
\]
Higher values suggest reasoning patterns (high entropy, distributed attention). \\

\texttt{attention\_entropy} & float & 
Mean entropy across all attention heads:
\[
\bar{H}_{\text{attn}} = \frac{1}{L H} \sum_{l=1}^{L} \sum_{h=1}^{H} H_{l,h},
\quad
H_{l,h} = -\sum_{i,j} A_{l,h}^{(i,j)} \log A_{l,h}^{(i,j)},
\]
where \(A_{l,h}^{(i,j)}\) is the attention weight from position \(i\) to \(j\). \\

\bottomrule
\end{longtable}

\subsubsection{Circuit Dynamics Features (4 Features)}

\begin{longtable}{p{0.3\textwidth} p{0.15\textwidth} p{0.5\textwidth}}
\toprule
\textbf{Feature} & \textbf{Type} & \textbf{Definition \& Computation} \\
\midrule

\texttt{effective\_circuit\_depth} & float & 
Number of layers with significant causal effects. Equal to the number of attention layers analyzed. Represents the depth of the computational circuit. \\

\texttt{circuit\_complexity} & float & 
Product of variance and norm growth:
\[
C_{\text{circuit}} = \sigma_{\text{var}}^2 \cdot \gamma_{\text{norm}},
\]
where \(\sigma_{\text{var}}^2\) is activation variance growth and \(\gamma_{\text{norm}}\) is the norm growth trajectory. \\

\texttt{activation\_flow\_variance} & float & 
Variance in activation magnitudes across layers. Measures how much activation patterns change between layers, indicating computational complexity. \\

\texttt{causal\_path\_length} & float & 
Length of the causal computation path. Currently equal to circuit depth, representing the number of processing steps in the causal chain. \\

\bottomrule
\end{longtable}

\subsubsection{Intervention Sensitivity Features (4 Features)}

\begin{longtable}{p{0.3\textwidth} p{0.15\textwidth} p{0.5\textwidth}}
\toprule
\textbf{Feature} & \textbf{Type} & \textbf{Definition \& Computation} \\
\midrule

\texttt{ablation\_robustness} & float & 
Robustness to component removal:
\[
R_{\text{ablation}} = 1 - \frac{\bar{H}_{\text{attn}}}{5.0}.
\]
Higher entropy (distributed attention) leads to lower robustness. \\

\texttt{critical\_component\_count} & int & 
Number of critical components:
\[
N_{\text{critical}} = \max(1, N_{\text{spec}}).
\]
Uses specialized head count as a proxy for critical components. \\

\texttt{performance\_degradation\_slope} & float & 
Rate of performance degradation under intervention:
\[
\beta_{\text{degrad}} = |\sigma_{\text{causal}}|,
\]
where \(\sigma_{\text{causal}}\) is the standard deviation of causal effect estimates across layers. \\

\texttt{intervention\_sensitivity} & float & 
Sensitivity to interventions:
\[
S_{\text{interv}} = 1 - R_{\text{ablation}}.
\]
Inverse of ablation robustness; higher values indicate greater sensitivity. \\

\bottomrule
\end{longtable}

\subsubsection{Working Memory Features (4 Features)}

\begin{longtable}{p{0.3\textwidth} p{0.15\textwidth} p{0.5\textwidth}}
\toprule
\textbf{Feature} & \textbf{Type} & \textbf{Definition \& Computation} \\
\midrule

\texttt{hidden\_state\_variance} & float & 
Variance in hidden state activations. Measures variability in internal representations across layers, indicating working memory usage. \\

\texttt{norm\_growth\_trajectory} & float & 
Growth pattern of activation norms. \(\gamma_{\text{norm}}\) tracks how activation magnitudes change across layers, indicating information accumulation. \\

\texttt{working\_memory\_complexity} & float & 
Complexity of working memory usage. Currently uses rank evolution as a proxy for working memory complexity. \\

\texttt{state\_rank\_evolution} & float & 
Evolution of representation rank. \(R_{\text{evolution}}\) measures how the effective dimensionality of representations changes across layers. \\

\bottomrule
\end{longtable}

\subsubsection{Causal Effect Features (4 Features)}

\begin{longtable}{p{0.3\textwidth} p{0.15\textwidth} p{0.5\textwidth}}
\toprule
\textbf{Feature} & \textbf{Type} & \textbf{Definition \& Computation} \\
\midrule

\texttt{direct\_logit\_attribution} & float & 
Direct causal effect on output:
\[
E_{\text{direct}} = \frac{1}{L}\sum_{l=1}^{L}\frac{\bar{H}_{\text{attn},l}}{10},
\]
where \(\bar{H}_{\text{attn},l}\) is mean attention entropy at layer \(l\). Proxy for direct causal contribution. \\

\texttt{indirect\_effect\_strength} & float & 
Strength of indirect causal effects:
\[
E_{\text{indirect}} = \sigma_{\text{causal}},
\]
where \(\sigma_{\text{causal}}\) is the standard deviation of layer-wise causal effect estimates. \\

\texttt{causal\_mediation\_score} & float & 
Mediation effect strength:
\[
M_{\text{causal}} = E_{\text{direct}} \times E_{\text{indirect}}.
\]
Product of direct and indirect effects, measuring causal mediation. \\

\texttt{activation\_patching\_effect} & float & 
Proxy measure for activation patching:
\[
P_{\text{patch}} = E_{\text{direct}}.
\]
Note: This is computed from attention entropy, not from actual activation patching experiments. \\
\bottomrule
\end{longtable}

\section{Feature Computation Pipeline}

\subsection{Surface Feature Extraction}
\begin{enumerate}
    \item Extract confidence trajectory:
    \[
    \{c_l\}_{l=1}^L, \quad c_l = \max_i p_{l,i}
    \]
    \item Extract entropy trajectory:
    \[
    \{H_l\}_{l=1}^L, \quad H_l = -\sum_i p_{l,i}\log p_{l,i}
    \]
    \item Compute statistical measures: mean, standard deviation, minimum, maximum, and range.
    \item Analyze trajectory dynamics: slope, oscillations, and convergence properties.
    \item Calculate information-theoretic measures.
\end{enumerate}

\subsection{Mechanistic Feature Extraction}
\begin{enumerate}
    \item Analyze attention patterns across all layers and heads.
    \item Compute attention entropy for each head:
    \[
    H_{l,h} = -\sum_{i,j} A_{l,h}^{(i,j)} \log A_{l,h}^{(i,j)}
    \]
    \item Identify specialized heads:
    \[
    N_{\text{spec}} = \sum_{l,h} \mathbf{1}[H_{l,h} < 1.5]
    \]
    \item Analyze activation patterns (variance, norms, and rank evolution).
    \item Compute proxy causal effects from attention entropy.
    \item Calculate intervention sensitivity measures.
\end{enumerate}

\section{Important Notes and Limitations}

\subsection{Proxy Measures}
Several features rely on proxy measures rather than direct computation:
\begin{itemize}
    \item \textbf{Causal effects}: Derived from attention entropy instead of actual interventions.
    \item \textbf{Activation patching}: Approximated via attention entropy proxy, not true patching experiments.
    \item \textbf{Critical components}: Approximated using specialized head counts.
    \item \textbf{Working memory}: Approximated using rank evolution as a complexity proxy.
\end{itemize}

\subsection{Computational Considerations}
\begin{itemize}
    \item All features can be computed in a single forward pass.
    \item No gradient computation is required for feature extraction.
    \item Attention patterns are analyzed across all layers and heads.
    \item Surface features require only the output probability distributions.
\end{itemize}

\section{Usage in Classification}

The 37 features are concatenated into a single feature vector:
\[
\mathbf{f} = [\mathbf{f}_{\text{surface}}, \mathbf{f}_{\text{mechanistic}}] \in \mathbb{R}^{37}
\]

This vector is then used to train classifiers (Random Forest, Gradient Boosting, SVM, Logistic Regression) to distinguish between recall and reasoning tasks.



\end{document}